\documentclass{article}

\usepackage{arxiv}

\usepackage[utf8]{inputenc} 
\usepackage[T1]{fontenc}    
\usepackage{hyperref}       
\usepackage{url}            
\usepackage{booktabs}       
\usepackage{amsfonts}       
\usepackage{nicefrac}       
\usepackage{microtype}      

\usepackage{amsmath}
\usepackage{graphicx}
\usepackage{subcaption}
\usepackage{biblatex}
\usepackage[colorinlistoftodos]{todonotes} 
\usepackage{soul}

\title{Realistic text replacement with non-uniform style conditioning}

\author{
    Arseny Nerinovsky\\
    Computer Technologies Laboratory\\
    ITMO University, St. Petersburg, Russia\\
    \texttt{nerinovsky.arseny@gmail.com}\\
\And
    Igor Buzhinsky\\
    Computer Technologies Laboratory, \\
    ITMO University, St. Petersburg, Russia \\
    \texttt{igor.buzhinsky@gmail.com}\\
\And
    Andey Filchencov \\
    Computer Technologies Laboratory\\
    ITMO University, St. Petersburg, Russia\\
    \texttt{afilchenkov@itmo.ru}\\
}

\newif\ifcomments
\commentstrue

\newcommand{\IB}[1]{\textcolor{orange}{#1}}
\newcommand{\soutIB}[1]{\IB{\st{#1}}}

\begin{document}
\maketitle

\begin{abstract}
In this work, we study the possibility of realistic text replacement,  the goal of which is to replace text present in the image with user-supplied text. The replacement should be performed in a way that will not allow  distinguishing the resulting image from the original one. We achieve this goal by developing a novel non-uniform style conditioning layer and apply it to an encoder-decoder ResNet based architecture. The resulting model is a single-stage model, with no post-processing. The proposed model achieves realistic text replacement and  outperforms existing approaches on ICDAR MLT.
\end{abstract}

\keywords{Text replacement\and GAN \and Style conditioning}

\section{Introduction}



\begin{figure}[h]

\setlength{\fboxsep}{0pt}%

\centering
\begin{subfigure}{.47\textwidth}
  \centering
  \fbox{\includegraphics[width=\linewidth,totalheight=2.5cm]{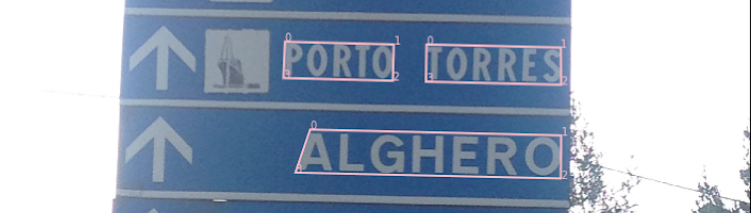}}
  \caption{Example of source image}
  \label{sub-fig-src}
\end{subfigure}
\begin{subfigure}{.47\textwidth}
  \centering
  \fbox{\includegraphics[width=\linewidth,totalheight=2.5cm]{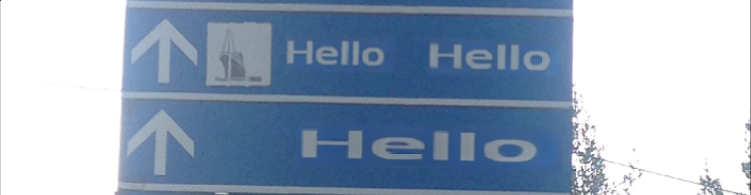}}
  \caption{Text replacement result}
  \label{sub-fig-tgt}
\end{subfigure}
\caption{Illustration of the text replacement task} 
\label{fig:form-ex}
\end{figure}

The task of realistic text replacement could be formulated as follows: replace text present in an image with arbitrary user-supplied text in a way that will not allow distinguishing the resulting image from the original one. This task is quite challenging since text is usually present in a variety of styles on a variety of backgrounds. An illustration of the text replacement task could be seen in Figure~\ref{fig:form-ex}, where all selected text (denoted by pink polygons) in the source image is substituted with the string ``hello''. 

We solve the task of realistic text replacement with a generative adversarial network (GAN)~\cite{goodfellow-gan} based on paper~\cite{pix2pix}. The generator is composed of a ResNet~\cite{res-nets} based encoder-decoder architecture. Text replacement is made with one forward pass trough the network without post-processing. In order to perform text replacement, two images are required---a \textit{content image}, which parametrizes what is inpainted, and a  \textit{style image}, which parametrizes the style of the inpainted images. Also masks denoting regions of the image where text is present are required. The parametrization of the inpainted areas is performed by replacing areas with images of text in the content image with areas filled with text edges. During training, the text edges correspond to the edges of the original text. During inference, the text edges correspond to the arbitrary user-supplied text. The style image is always equal to the source image. Examples of images could be seen in in Figure~\ref{fig:res-blk-arch}.


We address the challenge of different text styles by introducing a novel non-uniform conditioning layer called PatchedAdaIn. PatchedAdaIn allows us to extract user-delimited areas of style information from the style image and apply it to areas of the content image. This allows us to stylize different text instances present in the same image with different styles. The major contributions of this paper are:
\begin{itemize}
    \item \emph{PatchedAdaIn}, a non-uniform conditional normalization layer which allows applying different styles to different parts;
    \item a network achieving realistic text replacement, which we call \emph{Patched-Style GAN} (PsGAN). 
\end{itemize}




This paper is structured as follows. In Section~\ref{sec:related-work}, we describe works related to the proposed approach. Then, in Section~\ref{sec:proposed-approach}, we describe the proposed approach, we describe the non-uniform conditioning layer in Section~\ref{subsec:PatchedAdaIn} and the adopted architecture in Section~\ref{subsec:Architecture}. Then we describe the experiments performed in Section~\ref{sec:experiments}. Section~\ref{sec:conclusion} concludes the paper.

\section{Related work}
\label{sec:related-work}
\subsection{Generative adversarial networks}

GANs~\cite{goodfellow-gan} are a machine learning framework consisting of typically two networks, a generator network and a discriminator network. The task of the generator is to create samples which best resemble samples from the training dataset. The discriminator is tasked with differentiating generated and real samples. Both networks are trained together. 

In the original GAN formulation, the generator network maps samples from a normal distribution to images. The work of~\cite{cgans} introduces conditional GANs (cGANs). cGANs allow controlling their output by supplying a class label. Several improvements to the cGANs model are proposed in works~\cite{proj-d,inv-cgans}. They mainly differ in how the class label is supplied to the network. Several works explored the conditioning of GANs not only bound to a single class label. Variants of conditioning by text~\cite{att-gan}, bounding boxes and key-points~\cite{what-and-where-to-draw}, or images~\cite{pix2pix} exist. The latter is the most relevant to this work.


To allow image conditioning, the authors of~\cite{pix2pix} propose to use a Unet~\cite{unet} network for the generator, the proposed model is called Pix2Pix. The network maps a source image to a target image. Examples of source-target image pairs are segmentation maps and images from which such segmentation maps originate, black and white images and color variants of the former, and images with absent regions and the original images. The authors successfully apply the Pix2Pix network to a variety of source-target pairs. They refer to Pix2Pix as to a model for paired image translation due to its generality.



\subsection{Conditioning by modulation}

Batch normalization (BN) is an important part of many state-of-the-art neural network architectures. For example, the authors of DCGAN~\cite{dc-gan}, which is the first GAN that successfully applied a convolutional encoder and decoder, partly ascribe their success to aggressive use of BN. BN is composed of a normalization step followed by a modulation step.
During the normalization step, the input is normalized by substracting the mean $\mu_{b,h,w}$ and then dividing by the standard deviation $\sigma_{b,h,w}$:
$$\hat{y}(x) = \frac{x - \mu_{b,h,w}(x)}{\sigma_{b,h,w}(x)},$$
where the mean and standard deviation are computed across the batch, height and width dimensions. The modulation step is defined as $$\text{BN}(x) = \gamma \  \hat{y}(x) + \beta,$$
where $\gamma$ and $\beta$ are learned scale and shift parameters. 

Several variants of BN have been developed~\cite{layer-norm, group-norm}. The variation of BN most relevant to this work is instance normalization. First introduced in~\cite{instance-norm}, instance normalization was developed for aiding style transfer tasks. The definition of instance normalization differs from BN only in the way the statistics are computed. Unlike BN, instance normalization computes statistics across the channel dimension:
\begin{equation}
    \text{In}(x) = \gamma \frac{x - \mu_{h,w}(x)}{\sigma_{h,w}(x)} + \beta,
\end{equation}
where $\mu_{h,w}$ and $\sigma_{h,w}$ are the mean and standard deviation computed across the channel dimension. 

Conditional instance normalization is an extension of instance normalization proposed by~\cite{c-in}. The authors of~\cite{c-in} substitute the modulation parameters $\gamma$ and $\beta$ by  $\gamma^s$ and $\beta^s$ which are parametrized by the style image $s$. This approach is further improved by AdaIn~\cite{ada-in}, where instead of the learned parameters feature statistics are used. The latter allows avoiding pre-learned parameters. The AdaIn layer is formulated as 
\begin{equation}
    \text{AdaIn}(x, y) = \sigma_{h,w}(y)  \frac{x - \mu_{h, w}(x)}{\sigma_{h, w}(x)} + \mu_{h,w}(x).
\end{equation}

Conditional modulation was also later applied with different tweaks in many major GAN works, such as \cite{big-gan, style-gan1, style-gan2}. In general, the approach of conditional modulation of the network features has quite a varied applicability ranging from visual question answering~\cite{mod-vqa} to domain adaptation~\cite{li2018adaptive}. Conditional modulation is formalized in~\cite{Film} and given a more detailed treatment in~\cite{film}.




\subsection{Text editing}




Several works in the area of image generation are connected with the generation of text images. The most related to this paper is the work~\cite{editing-text}. The authors of~\cite{editing-text} generate realistic text replacements. Their proposed architecture takes as input a style image containing arbitrary text in a user-defined style and an input text image containing user-supplied text in a predefined style. Each image contains solely the stylized text. 

The task is split between three modules: a text conversion module, a background inpainting module and a fusion module. All three modules are learned independently. The text conversion module converts an image containing source text to an image where the source text has the correct color and style but no background. The background inpainting module is tasked with erasing text from the style image, leaving only the background. The task of the fusion module is to fuse the conversion module results and the background inpainting module results.



Another work related to image generation and text is~\cite{gen-text-seq}. Its main focus is the augmentation of datasets for text recognition. The authors of~\cite{gen-text-seq} propose an adaptation of the Pix2Pix network to realistically colour text images. The proposed architecture is composed of a combination of two Pix2Pix generators. The task of the first generator is to generate a coloured version of the text area and its surroundings. The task of the second generator is to fill the remaining background.


\section{Proposed approach}
\label{sec:proposed-approach}

To achieve realistic text replacement, we use an extension of the paired image translation paradigm introduced in~\cite{pix2pix}. The paired image translation paradigm could be formalized as: given a collection of source $\{ A_i \}_{i=0}^N$ and target $\{ T_i \}_{i=0}^N$ images, find a model $F$ capable of mapping source images to target ones: $\forall i\ F(A_i) = T_i$. 

Our proposed task formulation is a specialized extension of the paired image translation paradigm. We constrain the source image to images with text regions substituted by edge maps. We refer to the source image as to the \textit{content image} since the content of the text present in the generated images is expected to be equal to the content of the text in the content image. We also introduce one more input image---the \textit{style image}. The latter parametrizes the style of the inpainted regions---the inpainted text is expected to have the same background and foreground colour as the text present in the style image.  To create the content image we use an edge detection algorithm. The use of edges has the benefit of relieving us from the development of a specialized dataset.

\begin{figure}
  \centering
  \includegraphics[width=.8\linewidth]{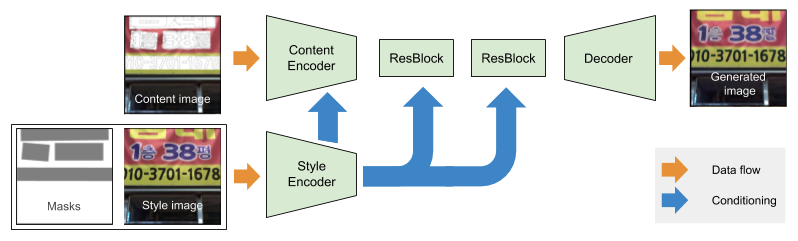}
  \caption{Scheme of the proposed architecture}
  \label{fig:res-blk-arch}
\end{figure}


The resulting expected mapping is $F' : \forall i\ F'(A_i, S_i, M_i) = T_i$, where $M_i$ is the mask image and $S_i$ is the style image. During training the style image is equal to the target image. The masks image $M_i$ is composed of masks denoting regions of text and is needed for technical reasons. Examples of content, style, target and mask images could be seen in Figure~\ref{fig:res-blk-arch}. 

\subsection{Non-uniform conditional normalization}
\label{subsec:PatchedAdaIn}

Unlike earlier works, we design an architecture capable of text replacement of several text areas in one forward pass. To accomplish this we introduce a novel non-uniform conditional normalization layer. Our design is based on the AdaIn layer. Like AdaIn, we demodulate features by subtracting the mean and dividing by the variance:
$$c(x) = \frac{x - \mu_{h, w}(x)}{\sigma_{h, w}(x)}.$$

We adapt the layer modulation by introducing an additional parameter---\textit{style patches} areas of the image having a uniform style. Like AdaIn, we use statistics of a style feature to modulate the demodulated layer inputs. We allow such modulation to be performed not on the whole image but on predefined areas delimited by the style patches parameter. 

We iterate through all the patches and extract mean and standard deviation statistics of the areas denoted by the patches from the style features\footnote{We do not directly iterate through the masks, but use PyTorch broadcasting. Although the implementation still requires a loop though the batch dimension, the resulting code has a negligible performance penalty.}. Both statistics are calculated across the width and height dimension. Then we\soutIB{,} subtract the mean and divide by the standard deviation, but do so only in the region denoted by the mask\IB{:}
\begin{subequations}
\begin{align}
r_\text{part} &= \sum_{i=1}^M (c(x) \odot m_i) \odot \sigma_{h, w}(s \odot m_i) + \mu_{h, w}(s \odot m_i),
\end{align}
\end{subequations}
where $M$ is the number of style patches and $\odot$ is element-wise multiplication. As a final step, we modulate all the space not covered by any masks. We refer to such space as to the \textit{background}. We perform modulation in a similar fashion: extract the mean and standard deviation and respectively subtract and divide the content features by them, and the statistics are extracted from the style image:
\begin{subequations}
\begin{align}
m_\text{bg} &=  1 - \bigvee_{i=1}^{M} m_i, \\
r &= r_\text{part} + (c(x)\odot m_\text{bg}) \odot \sigma_{h, w}(s \odot m_\text{bg}) + \mu_{h, w}(s \odot m_\text{bg}),
\end{align}
\end{subequations}
where $r$ is the layer output. We call the proposed layer PatchedAdaIn.

\subsection{Architecture}
\label{subsec:Architecture}

The proposed architecture consists of an encoder, several residual blocks (ResBlocks), and a decoder. The scheme could be seen in Figure~\ref{fig:res-blk-arch}. Conditioning is performed in the encoder and in all the residual blocks. Conditioning in the encoder was performed on features extracted from the style encoder. The style branch is identical to the encoder of the main network with one residual block.

\begin{figure}[h]
  \centering
  \includegraphics[width=.9\linewidth]{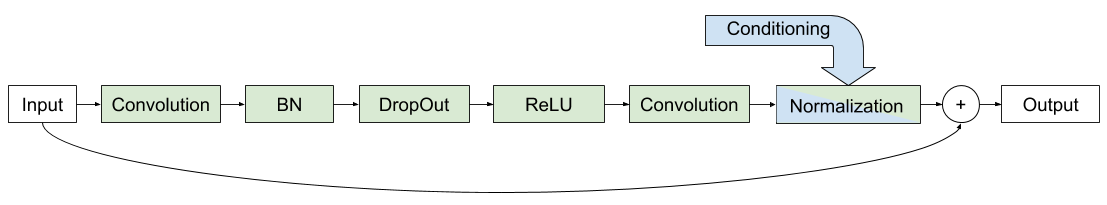}
  \caption{Scheme of a conditioned ResBlock}
  \label{fig:res-blk-blk}
\end{figure}

In order to allow conditioning, we changed the structure of the ResBlock. The default ResBlock is structured as a pair of consecutive convolutions followed by a normalization layer. The convolutions are separated by a ReLU and a dropout layer. The modified conditioned residual blocks differed form their regular counterparts by the last normalization layer, which was substituted with a conditional PatchedAdaIn normalization layer. A scheme of the proposed ResBlock could be seen in Figure~\ref{fig:res-blk-blk}.

\begin{figure}[h]
  \centering
  \begin{subfigure}{.24\textwidth}
    \centering
    \includegraphics[width=\linewidth]{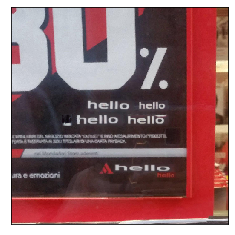}
  \end{subfigure}
  \begin{subfigure}{.24\textwidth}
    \centering
    \includegraphics[width=\linewidth]{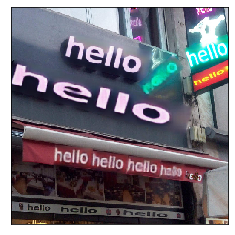}
  \end{subfigure}
  \begin{subfigure}{.24\textwidth}
    \centering
    \includegraphics[width=\linewidth]{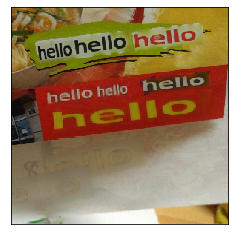}
  \end{subfigure}
  \begin{subfigure}{.24\textwidth}
    \centering
    \includegraphics[width=\linewidth]{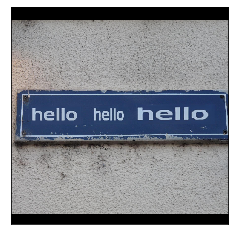}
  \end{subfigure}
    \begin{subfigure}{.24\textwidth}
    \centering
    \includegraphics[width=\linewidth]{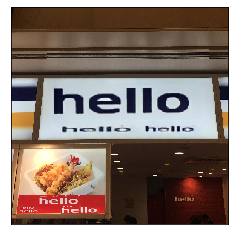}
  \end{subfigure}
  \begin{subfigure}{.24\textwidth}
    \centering
    \includegraphics[width=\linewidth]{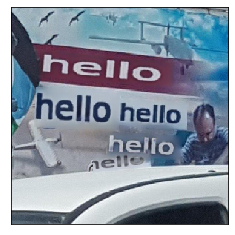}
  \end{subfigure}
  \begin{subfigure}{.24\textwidth}
    \centering
    \includegraphics[width=\linewidth]{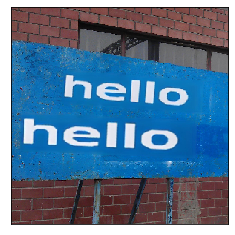}
  \end{subfigure}
  \begin{subfigure}{.24\textwidth}
    \centering
    \includegraphics[width=\linewidth]{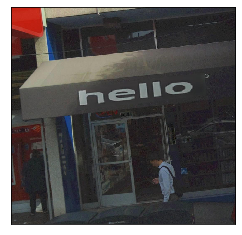}
  \end{subfigure}
    \begin{subfigure}{.24\textwidth}
    \centering
    \includegraphics[width=\linewidth]{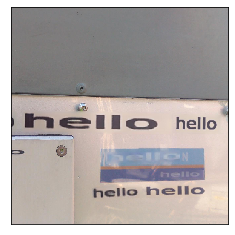}
  \end{subfigure}
  \begin{subfigure}{.24\textwidth}
    \centering
    \includegraphics[width=\linewidth]{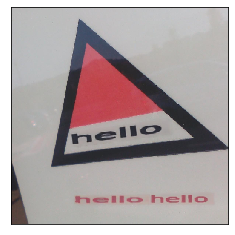}
  \end{subfigure}
  \begin{subfigure}{.24\textwidth}
    \centering
    \includegraphics[width=\linewidth]{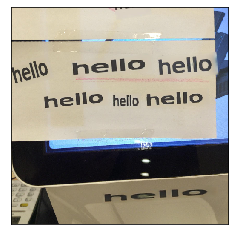}
  \end{subfigure}
  \begin{subfigure}{.24\textwidth}
    \centering
    \includegraphics[width=\linewidth]{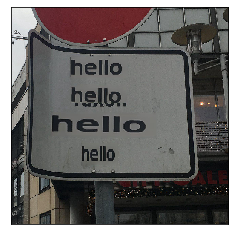}
  \end{subfigure}
  
  \caption{Text inpainting examples } 
  \label{fig:succ-res-blk}
\end{figure}

The training is performed with $L_1$ regularization and adversarial losses as described in the Pix2Pix paper~\cite{pix2pix}. Several successful examples of the model could be seen in Figure~\ref{fig:succ-res-blk}.

\section{Experiments}
\label{sec:experiments}

\subsection{Data}

\begin{figure}
\centering
\begin{subfigure}{.243\textwidth}
  \includegraphics[width=4cm,totalheight=2.5cm]{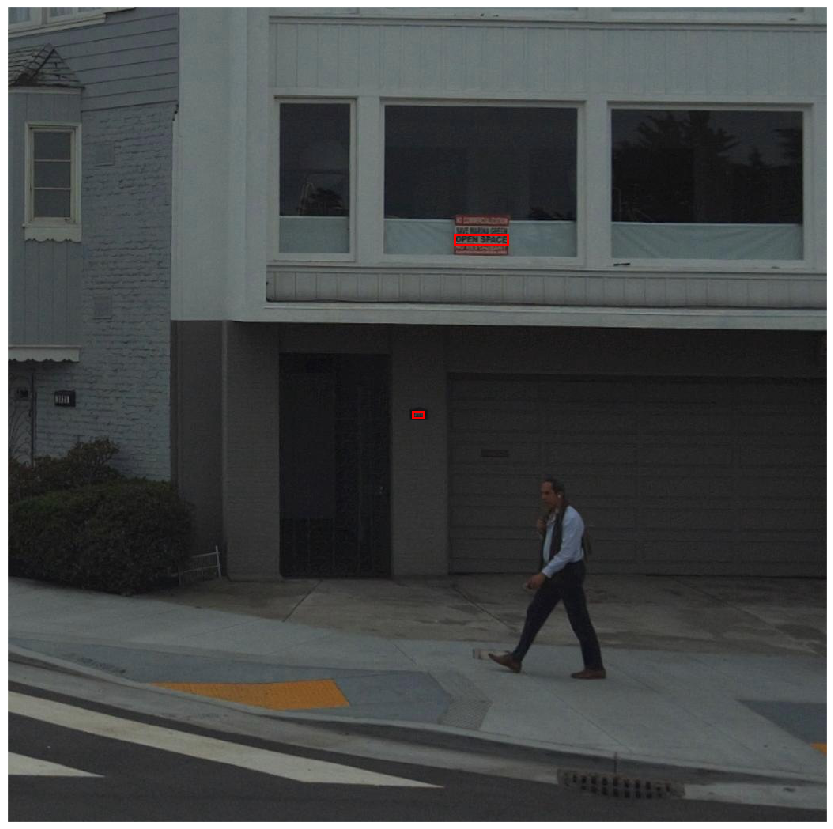}
\end{subfigure}
\begin{subfigure}{.243\textwidth}
  \includegraphics[width=4cm,totalheight=2.5cm]{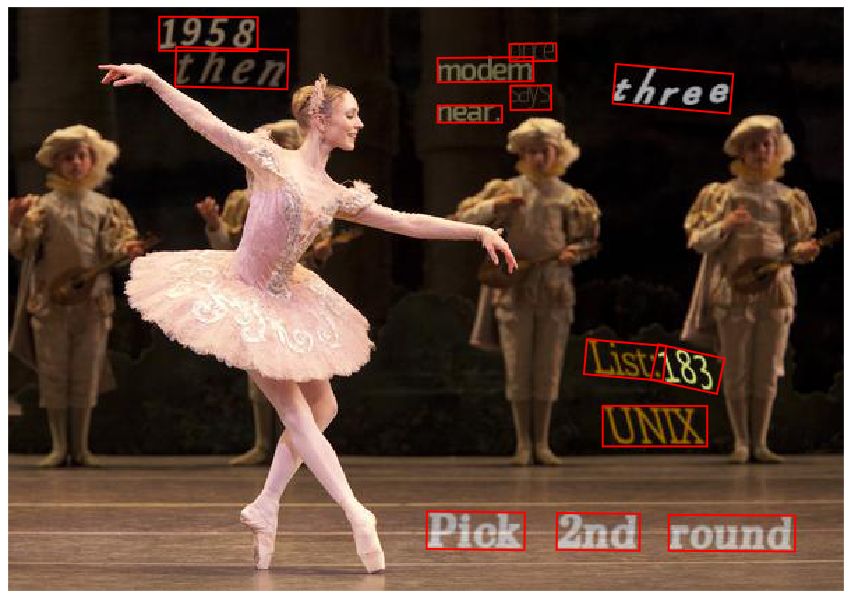}
\end{subfigure}
\begin{subfigure}{.243\textwidth}
    \includegraphics[width=4cm,totalheight=2.5cm]{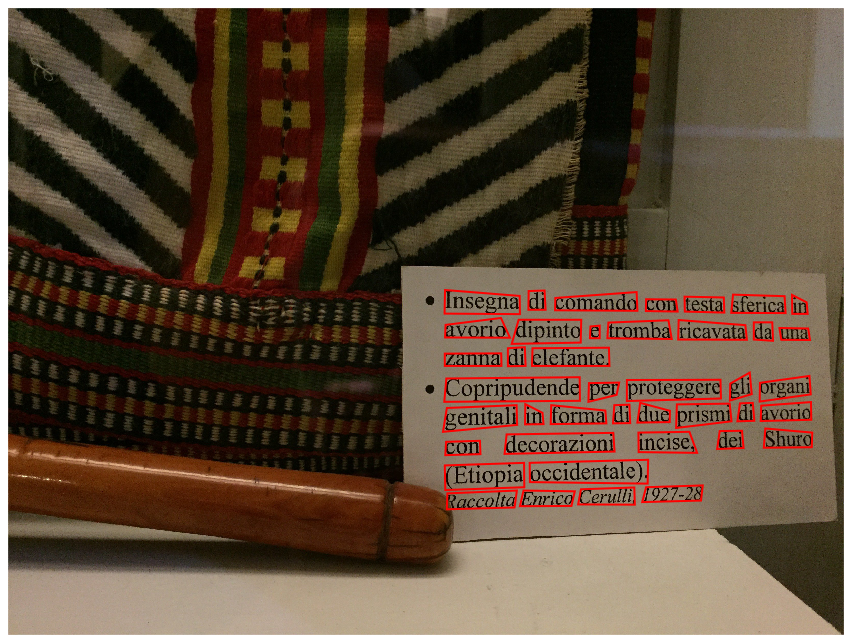}
\end{subfigure}
\begin{subfigure}{.243\textwidth}
  \includegraphics[trim={0 5cm 0 1cm}, clip,width=4cm,totalheight=2.4cm]{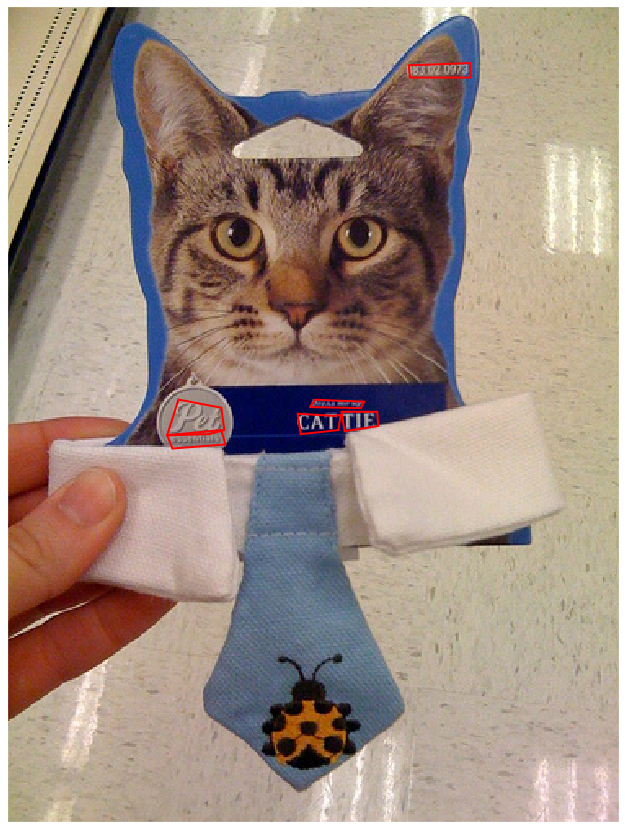}
\end{subfigure}
\caption{Samples from text detection datasets} 
\label{fig:dataset-samples}
\end{figure}

We used several text detection datasets: ICDAR MLT~\cite{mlt}, COCO Text~\cite{COCO-Text}, UberText~\cite{UberText} and SynthText~\cite{synth-text}. All datasets contain images and polygons delimiting areas where text is present. Several samples could be seen in Figure~\ref{fig:dataset-samples}.

We estimated the approximate size of the dataset required based on the numbers required by sketch-to-image models and by image inpainting models. Our findings suggested that a dataset in the range of 100,000 samples 
will be enough, since Pix2Pix was trained as a sketch to image model on 137 thousand image pairs from the Amazon Handbag images dataset~\cite{igan}. Context encoders were successfully trained on 100 thousand images form 100K-Imagenet and \cite{inp-convs} were successfully trained on 27 thousand images from CELEBA-HQ~\cite{maskgan} and 50 thousand inpainting masks. 

To generate data for training, we used data from text detection datasets. One peculiarity of the data is that the distribution of polygon sizes is skewed towards smaller polygons. The polygons could be so small that the text delimited by them becomes invisible. As the input to the network, we should supply a square image with reasonably sized and reasonably dense text. The nature of the data makes supplying every text instance to the network unfeasible. Thus, we rejected polygons whose largest dimension spans under 200 pixels in length. The total size of the agglomerate dataset is 245,225 samples.

\subsection{Metrics}

To assess the quality of the resulting images, we measured the validation score of a pre-trained model for text recognition. A good validation performance is indicative of good resulting images. Word error rate (WER) and accuracy were estimated. We used ICDAR MLT as our validation dataset.  We calculated such metric by using the CRNN model from~\cite{crnn}. We also calculated the accuracy and word error rate metric. These metrics were calculated both on a per dataset basis and on a per-image basis. The difference between the two is that the first is the average of all the resulting metrics while the second is the average of metrics averaged by image. We call this set of metrics \textit{validation proxy metrics}.

Another way to assess the performance of the inpainting model is to enhance the training set of a text recognition model with data generated by the inpainting model. An increase in the training score will be indicative of good inpainting model. We call this set of metrics \textit{training proxy metrics}.

\subsection{Results}

The base architecture is currently capable of reaching a per image accuracy of 0.668, which is comparable to the accuracy of 0.824 achieved by the model on raw data. We outperform the work~\cite{editing-text}, which achieves a per image accuracy of 0.517. The values of the validation proxy metrics could be seen in Table~\ref{tbl:base}. 

\begin{table}
    \centering
    \caption{Validation metrics of a pretrained CRNN network}
    \renewcommand{\arraystretch}{1.3}
    \begin{tabular}{ lcccc }
    \hline Model & Accuracy &  Per image accuracy & WER & Per image WER \\
    \hline PsGAN (Ours) & 0.653 & 0.668 & 0.0708 & 0.0680 \\
    ICDAR MLT  & 0.813 & 0.824 & 0.0384 & 0.0323 \\
    Generated by~\cite{editing-text}  & 0.544 & 0.517 & 0.0901 & 0.1020 \\
\hline
    \end{tabular}
    \label{tbl:base}
\end{table}


In~\cite{gen-text-seq}, proxy metrics are trained to assess the model quality. The authors of~\cite{gen-text-seq}  train a model on 8 million synthetically generated images. In our case, because of the size of our network, such validation would be unfeasible. Instead, we augment the training dataset with additional synthetic images. The number of synthetic images is equal to 25\% of the number of original training images. The model trained on the dataset augmented with synthetic images achieves a  lower WER score. The training proxy metrics could be seen in Table~\ref{tbl:wer-plus}.

\begin{table}
    \centering
    \caption{Results of training a CRNN model on a dataset augmented with synthetic data}
    \renewcommand{\arraystretch}{1.3}
    \begin{tabular}{ lc } 
    \hline CRNN training dataset & WER \\
    \hline ICDAR MLT & 0.038 \\
            ICDAR MLT + 25\% synthetic text & 0.015 \\
    
\hline
    \end{tabular}
    \label{tbl:wer-plus}
\end{table}



\section{Conclusions}
\label{sec:conclusion}

This paper proposes a feedforward model for realistic text replacement based on a ResBlock encoder-decoder architecture. A novel non-uniform conditioning normalization layer is introduced to allow the application of different styles to different parts of the image. The proposed models outperform all previous works on all the measured metrics and achieve high image realism.


\section*{Acknowledgments}

The research was financially supported by the Government of the Russian Federation, Grant 08-08.

\printbibliography

\end{document}